\documentclass[preprint,12pt]{elsarticle}

\usepackage{amssymb}

\usepackage{amsmath}
\usepackage{booktabs, multirow}
\usepackage{tabularx}
\usepackage{setspace}
\usepackage{float}
\usepackage{subfig}

\begin{document}

\begin{frontmatter}

\title{3D Face Alignment Through Fusion of Head Pose Information and Features}

\author{Jaehyun So\fnref{1}}
\ead{ru2ror@soongsil.ac.kr}
\affiliation[1]{organization={Soongsil University},
            addressline={Department of Electronic Engineering},
            state={Seoul},
            country={Republic of Korea}}

\author{Youngjoon Han\corref{cor1}\fnref{2}}
\ead{young@ssu.ac.kr}
\affiliation[2]{organization={Soongsil University},
           addressline={School of AI Convergence}, 
           state={Seoul},
           country={Republic of Korea}}
\cortext[cor1]{Corresponding author}

\begin{abstract}
The ability of humans to infer head poses from face shapes, and vice versa, indicates a strong correlation between the two. Accordingly, recent studies on face alignment have employed head pose information to predict facial landmarks in computer vision tasks. In this study, we propose a novel method that employs head pose information to improve face alignment performance by fusing said information with the feature maps of a face alignment network, rather than simply using it to initialize facial landmarks. Furthermore, the proposed network structure performs robust face alignment through a dual-dimensional network using multidimensional features represented by 2D feature maps and a 3D heatmap. For effective dense face alignment, we also propose a prediction method for facial geometric landmarks through training based on knowledge distillation using predicted keypoints. We experimentally assessed the correlation between the predicted facial landmarks and head pose information, as well as variations in the accuracy of facial landmarks with respect to the quality of head pose information. In addition, we demonstrated the effectiveness of the proposed method through a competitive performance comparison with state-of-the-art methods on the AFLW2000-3D, AFLW, and BIWI datasets.
\end{abstract}


\end{frontmatter}

\section{Introduction}
Many methods have been studied to analyze facial attribute information for human-computer interactions. In particular, face alignment and head pose estimation have been associated with good performance in the tasks of predicting facial landmarks and face orientation, respectively, using deep learning algorithms in computer vision. 

Face shapes and head poses are known to be strongly correlated in the visual field. Based on learned prior knowledge, humans can easily infer a head pose by looking at a face shape; conversely, humans can also easily infer a face shape by considering the corresponding head pose. This indicates that head pose information is important for the task of face alignment. Conventional face alignment methods~\cite{cootes2000introduction, cootes2001active} employ face models based on principal component analysis (PCA) to reduce the dimensions of facial landmarks. In recent years, many studies on 3D face alignment have employed 3D morphable models (3DMMs)~\cite{blanz2003face}, with PCA models used to represent scale, rotation, translation, and PCA parameters. The scale and translation are normalized by a face bounding box, and the PCA parameters represent a slight offset from the mean landmarks, but the rotation parameters have a significant effect on the landmarks. A correlation between head pose information and facial landmarks can also be observed in studies on nonparametric face alignment. For example, Feng et al.~\cite{feng2018wing} demonstrated a similarity between the distribution of 2D facial landmarks projected onto a principal axis using PCA and the rotation of the head on the yaw axis.

\begin{figure}[!htb]
	\centering
		\includegraphics[width=135mm]{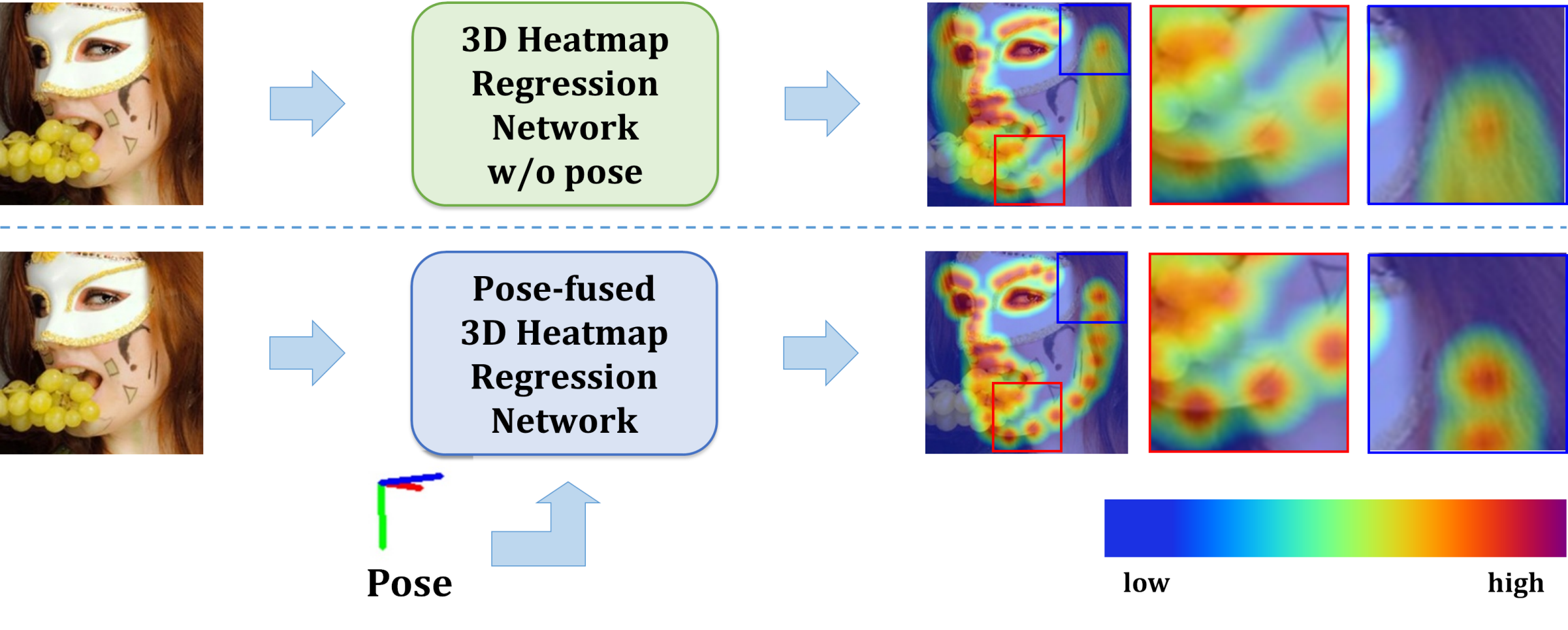}
	  \caption{Relationship between head pose information and 3D heatmap regression, reflected through differences between predicted 3D heatmaps. Warmer colors indicate a higher incidence of landmarks. Close-up images are shown of the red- and blue-bounded regions in the heatmaps.}
	  \label{fig:effectiveness_pose}
\end{figure}

However, previous studies have not sufficiently employed the correlation between head pose information and facial landmarks, instead using said information to initialize mean landmarks via rotation matrices and similar tools. To fill this research gap, we propose a novel approach called pose-fused face alignment (PFA) to improve face alignment performance. The PFA network initially predicts a pre-pose as head pose information, and subsequently fuses the pre-pose and feature maps to perform face alignment. The proposed method robustly predicts facial landmarks in an occluded environment using a facial profile. We used a vectorized 3×3 rotation matrix and channel attention~\cite{hu2018squeeze} to fuse the head pose information into feature maps extracted from the residual block of the PFA network. Figure~\ref{fig:effectiveness_pose} illustrates the differences between 3D heatmaps predicted with and without the pre-pose.

The proposed PFA network comprises a two-stage structure. The pre-stage consists of three subnetworks for the respective tasks of feature extraction, pre-pose regression, and 3D heatmap regression. The pre-pose regression network predicts the pre-pose using feature maps extracted by the feature extraction network, and propagates the pre-pose to the post-networks, whereas the 3D heatmap regression network predicts a 3D facial landmark heatmap. The post-stage consists of a dual-dimensional network that performs sparse face alignment by predicting keypoint landmarks using the pre-pose, 2D feature maps, and 3D heatmaps extracted in the pre-stage. In addition, this network includes a fully-connected layer that predicts geometric landmarks through training based on knowledge distillation using the predicted keypoints, referred to as keypoint-to-geometry (K2G). We also improved the performance of geometric landmark prediction to ensure alignment with the predicted keypoint landmarks referred to as geometry-to-keypoint (G2K). Our primary contributions can be summarized as follows:

\begin{enumerate}
\item A  network was developed to fuse head pose information into feature maps, yielding improved face alignment performance.
\item We designed a network structure for face alignment, as well as a dual-dimensional network composed of 2D and 3D kernels, to address the limitations of 3D heatmaps. 
\item We performed dense face alignment training based on knowledge distillation to align predicted geometric landmarks with predicted keypoint landmarks.
\item We achieved a reduction in computational cost with a slight decrease in performance using the methods described above.
\item The proposed PFA network achieves performance competitive with that of state-of-the-art methods. An ablation study was conducted to demonstrate the effect of head pose information on the accuracy of face alignment.
\end{enumerate}

\section{Related Work}
Face alignment is the task of predicting landmarks to represent facial components in an image. This process can be classified as sparse face alignment or dense face alignment according to the type of facial landmarks. Previous approaches in this field include parametric methods that employ PCA-based modeling, and non-parametric methods that directly predict the coordinates of facial landmarks.

\subsection{Parametric face alignment}
Parametric face alignment is an approach wherein low-dimensional face parameters are predicted by reducing the number of high-dimensional facial landmarks using PCA. Many researchers have studied the prediction of PCA parameters representing facial geometry by applying 3DMMs to images. Blanz et al.~\cite{blanz2003face} proposed a 3DMM that expresses information regarding the facial geometry. Zhu et al.~\cite{zhu2016face} and Guo et al.~\cite{guo2020towards} proposed training methods for dense face alignment using keypoint landmarks within geometric landmarks transformed from PCA parameters to coordinates. Wu et al.~\cite{wu2021synergy} improved the training method using predicted keypoint landmarks projected on the PCA space of a 3DMM. Li et al.~\cite{li2023dsfnet} fused facial geometries constructed using PCA parameters that were directly and indirectly predicted by four types of feature maps. The aforementioned 3DMM-based methods realize multi-task learning and generalizability by training the models on head pose information and face geometry simultaneously, enabling them to efficiently predict high-dimensional geometric landmarks in a lower dimensionality.

\subsection{Non-parametric face alignment}
Under the non-parametric approach, facial landmarks are predicted through a coordinate or heatmap regression without considering PCA parameters. Studies on coordinate regression, wherein the coordinates of facial landmarks are predicted directly, have primarily focused on processing high-dimensional coordinates. Feng et al.~\cite{feng2018joint} represented geometric landmarks using a 2D UV position map corresponding to 3D coordinates. Ruan et al.~\cite{ruan2021sadrnet} proposed a robust face alignment method that predicts a 2D UV position map and an attention mask for occlusion. In contrast, a heatmap regression predicts the probability of a landmark’s existence in each pixel of a heatmap. This approach was developed to overcome the challenge associated with representing 3D landmarks through heatmaps. Bulat et al.~\cite{bulat2017far} proposed a method that predicts the x- and y-coordinates of 3D keypoint landmarks through a 2D heatmap regression, and subsequently predicts the corresponding z-coordinates by applying coordinate regression to the predicted 2D heatmaps. Zhang et al.~\cite{zhang2019adversarial} introduced a method that predicts 3D keypoint landmarks using a compound 3D heatmap that unifies all 3D keypoint heatmaps in a single space. Prados et al.~\cite{Prados-Torreblanca_2022_BMVC} used head pose information predicted via parametric methods to initialize landmarks and iteratively predict landmark displacement for 2D face alignment. Xia et al.~\cite{xia2023robust} proposed a method that simultaneously predicts landmark coordinates and uncertainty using transformer structures designed in three stages using landmark patches. Although the aforementioned methods have been developed for the accurate prediction of landmarks, most of them fail to sufficiently consider head pose information.

\section{Proposed Method}

\begin{figure}[!htb]
	\centering
		\includegraphics[width=135mm]{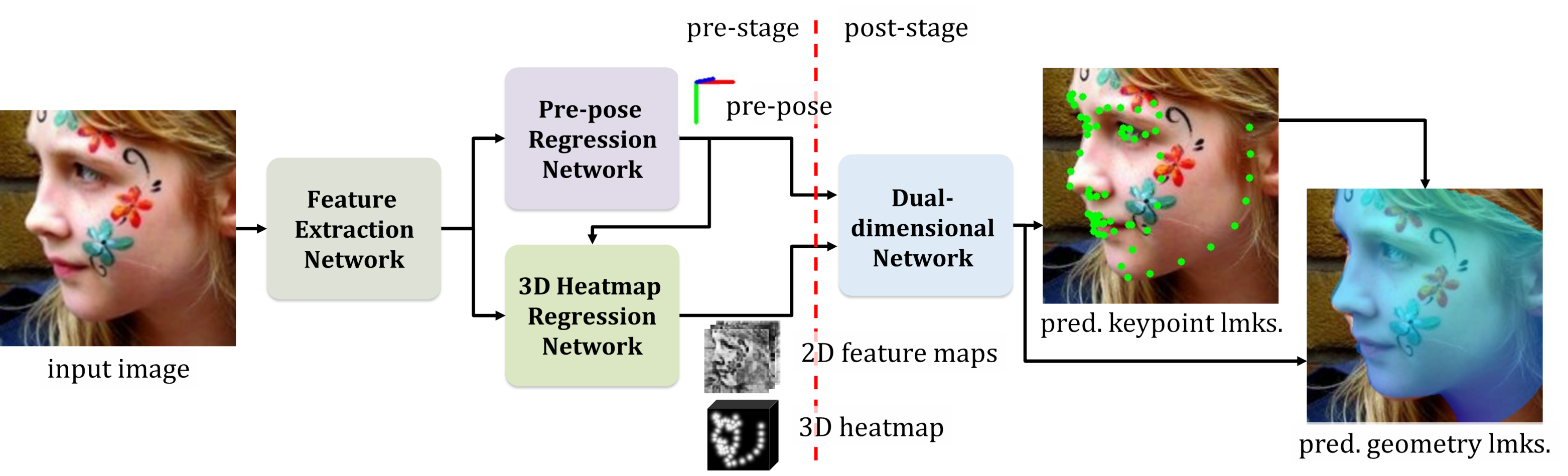}
	  \caption{Overview of pose-fused face alignment (PFA) network. A pre-pose is initially predicted in the pre-stage, and a 3D heatmap is subsequently predicted using said pre-pose. In the post-stage, facial landmarks are predicted using the predicted pre-pose, 2D feature maps, and 3D heatmap.}
	  \label{fig:network}
\end{figure}

The method proposed in this paper improves face alignment performance not only through multitask learning, but also by fusing head pose information into feature maps. Figure~\ref{fig:network} illustrates the PFA network structure. The full network structure is composed of four subnetworks – respectively used for feature extraction, pre-pose regression, and 3D heatmap regression – in the pre-stage, and a dual-dimensional subnetwork in the post-stage. The feature extraction network extracts feature maps from an input image, and propagates them to the pre-pose and 3D heatmap regression networks. The pre-pose regression network predicts a pre-pose as head pose information and propagates it to the 3D heatmap regression network, which predicts a 3D heatmap by fusing feature maps with the predicted pre-pose. The 2D feature maps and 3D heatmap extracted by this network are then propagated to the post-stage, wherein the dual-dimensional network uses this information to predict landmarks for face alignment; i.e., the network performs sparse face alignment. Following the training phase for sparse face alignment, a fully-connected layer is appended to the trained dual-dimensional network and trained for dense face alignment – i.e., the prediction of geometric landmarks – based on knowledge distillation using the predicted keypoint landmarks.

\subsection{Pose-fused residual block}
\label{sec:pfrb}

\begin{figure}[!htb]
	\centering
		\includegraphics[width=135mm]{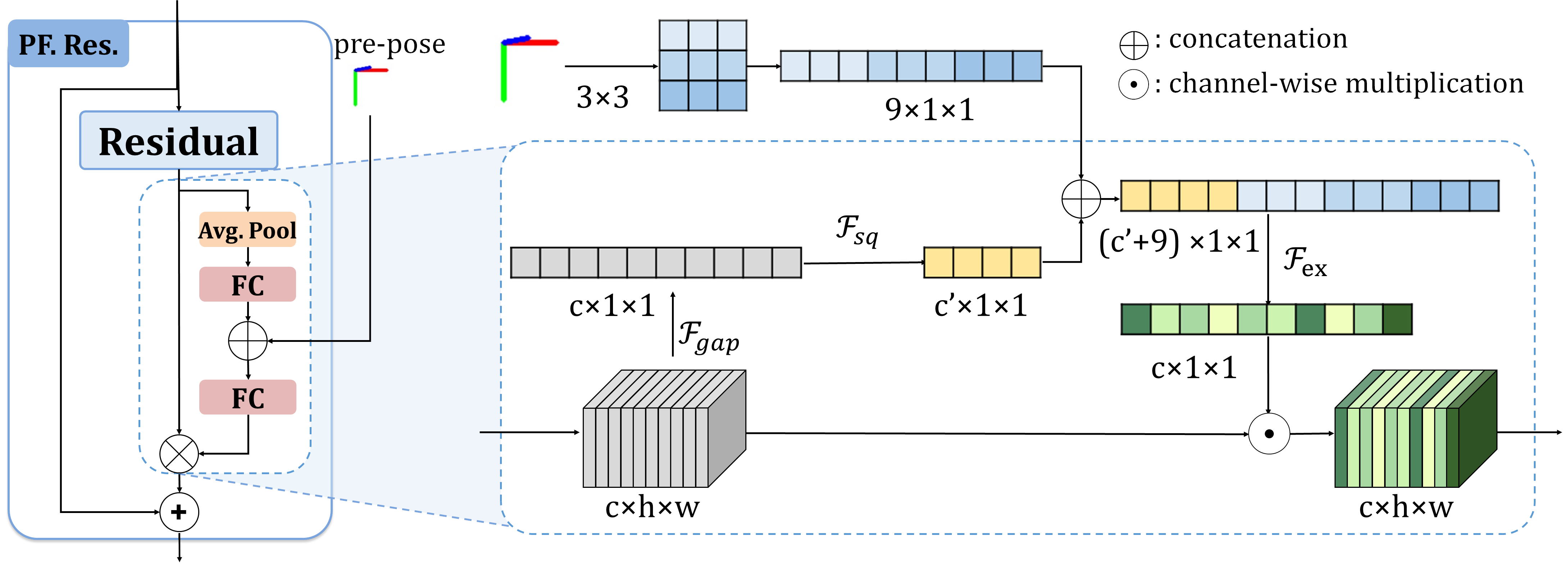}
	  \caption{Structure of pose-fused residual block (PFRB). A rotation matrix is concatenated with an encoded vector from the feature maps prior to rescaling said feature maps. c is the number of feature channels; c$'$ is the number of feature channels reduced by $\mathcal{F}_{sq}$; h and w are dimensions of the feature map.}
	  \label{fig:pfrb}
\end{figure}

Because the head pose represents one of the primary pieces of information representing the face, we improved face alignment performance by fusing head pose information into feature maps within a residual block called the pose-fused residual block (PFRB). This fusion process is performed through a channel attention method. Figure~\ref{fig:pfrb} illustrates the structure of the PFRB based on squeeze-and-excitation (SE)~\cite{hu2018squeeze}. Channel attention is performed using a weight vector calculated by concatenating the rotation matrix and encoded vector from the residual feature maps.

\begin{equation}
    F_{pg} = F_{res} \odot \mathcal{F}_{ex} (R \oplus (\mathcal{F}_{sq} \circ \mathcal{F}_{gap}(F_{res}))),
\label{eq:pose_fusion}
\end{equation}

\begin{figure}[!htb]
	\centering
		\includegraphics[width=135mm]{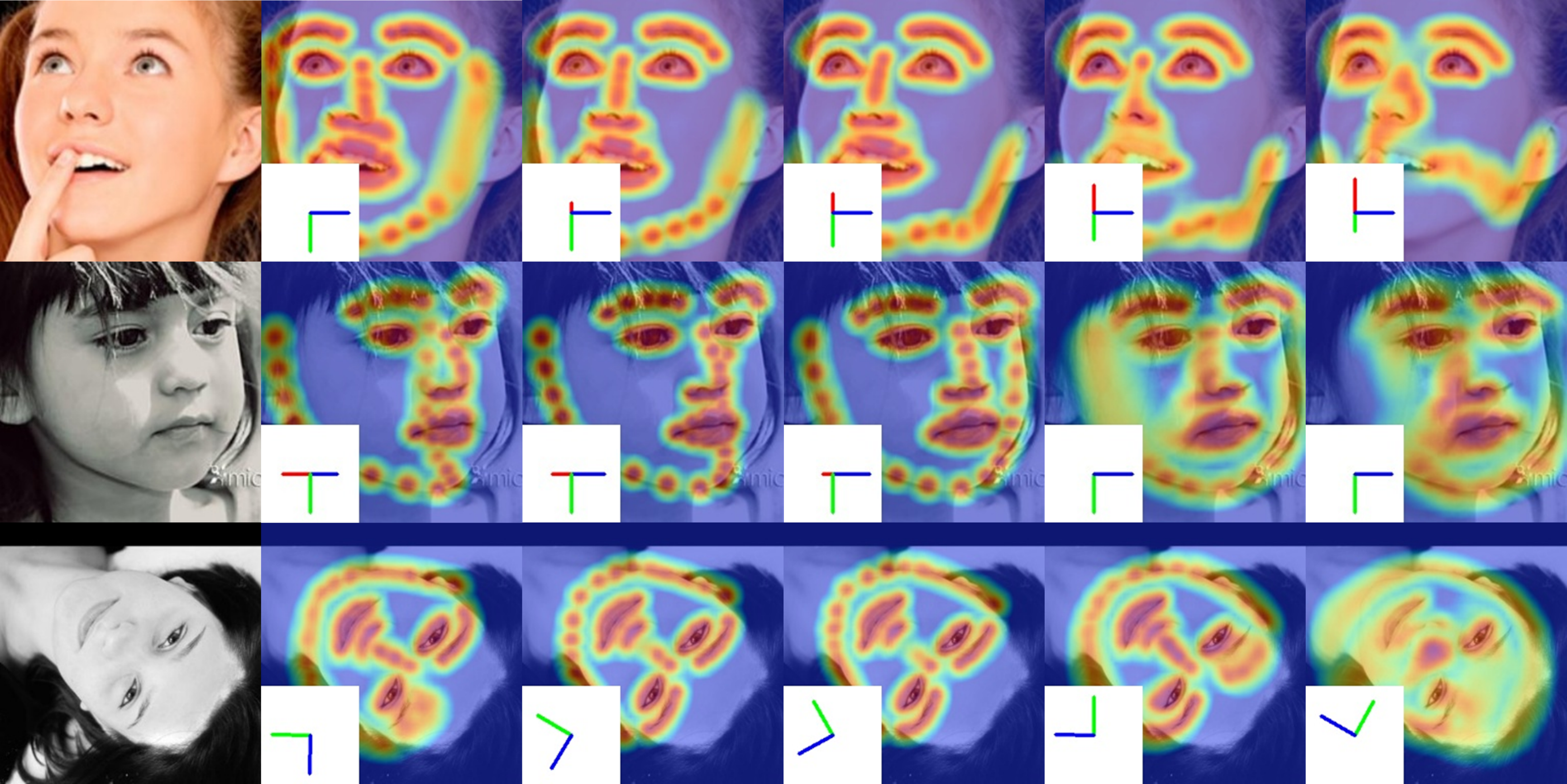}
	  \caption{Results of 3D heatmap regression fused with pre-defined head poses. These examples were predicted by fixing the two axes to 0° and subsequently shifting one axis. Examples in the first row reflect changes in the pitch angle, those in the second row reflect changes in the yaw angle, and those in the third row reflect changes in the roll angle. The orientation in the bottom-left corner of each heatmap denotes the pre-defined head pose. Pitch and yaw angles were changed by an interval of 10°, whereas the roll angle was changed by an interval of 20°.}
	  \label{fig:posetest}
\end{figure}

\noindent Equation~(\ref{eq:pose_fusion}) represents the fusion process, where $F_{pg}$ is a pose-fused feature map, $F_{res}$ is a residual feature map, $\odot$ denotes channel-wise multiplication, $\oplus$ denotes concatenation, $\mathcal{F}_{ex}$ is an excitation operation, $R$ is a vectorized rotation matrix, $\mathcal{F}_{sq}$ is a squeeze operation, and $\mathcal{F}_{gap}$ is a global average pooling operation. The distinguishing aspect of SE~\cite{hu2018squeeze} is that head pose information is fused into feature maps through $R$. In Figure~\ref{fig:pfrb}, c$’$ is the number of channels reduced by the ratio $r$: $\mathrm{c}’ = \mathrm{c} / r$. We set $r$ to 16 for the purpose of this study. Figure~\ref{fig:posetest} depicts 3D heatmaps predicted by fusing predefined head poses at uniform intervals to the 3D heatmap regression network trained using ground truth head pose information as a pre-pose.

\noindent The heatmaps shown in Figure~\ref{fig:posetest} were predicted differently based on the input head poses. When the input head pose information matched that of the image, the landmark heatmaps were distinct for the face components. In cases of a mismatch between the input and image, a heatmap was induced by the head pose while maximally predicting the facial components. These results demonstrate a strong correlation between head pose information and facial landmarks, with the former being a significant factor in predicting the latter.

\subsection{Pre-pose and 3D heatmap regression network}
\label{sec:prenet}

\begin{figure}[!htb]
	\centering
		\includegraphics[width=135mm]{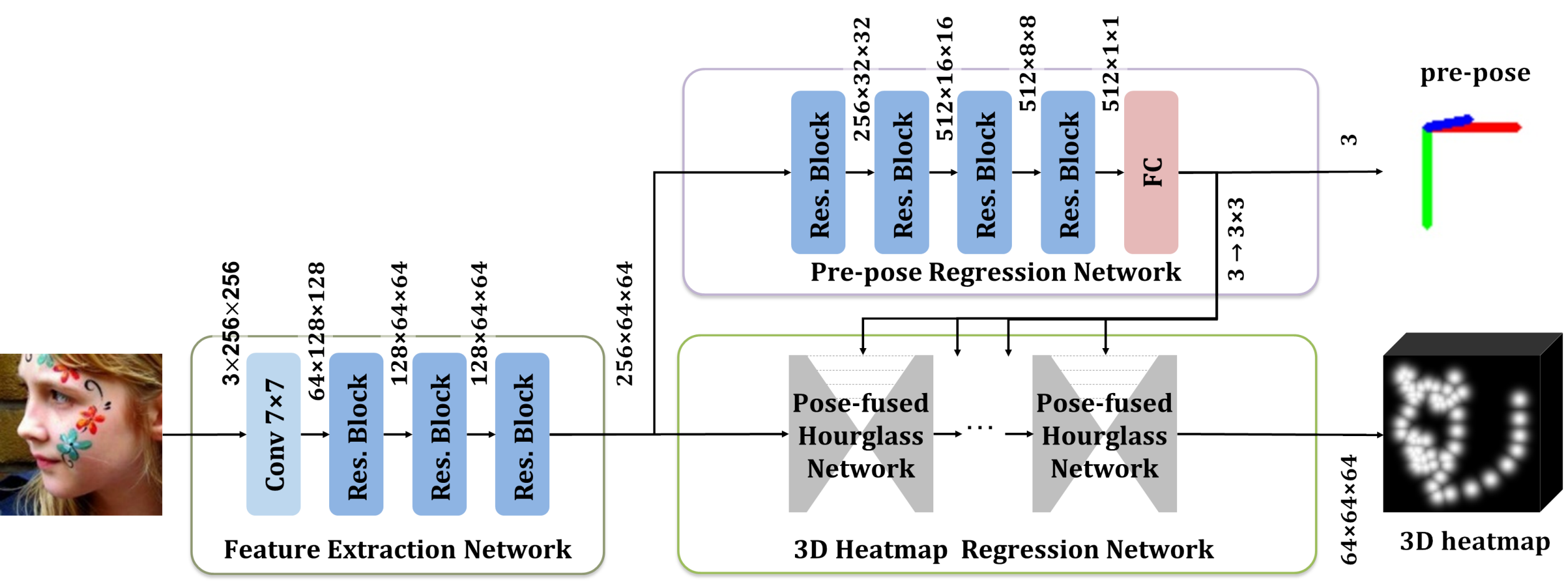}
	  \caption{Structure of pre-pose and 3D heatmap regression network.}
	  \label{fig:prestage}
\end{figure}

The pre-stage encompasses the extraction of the pre-pose, 2D feature maps, and 3D heatmap, all of which are propagated to the post-stage. The feature extraction network extracts feature maps and passes them to the pre-pose and 3D heatmap regression networks. The pre-pose regression network predicts a pre-pose through the four residual blocks, and propagates it to the 3D heatmap regression network, which predicts a 3D heatmap through four-stacked hourglass networks consisting of pose-fused residual blocks that fuse the predicted pre-pose into feature maps.

In heatmap regression, the occurrence of extraneous background pixels poses a challenge in network training. Because 3D heatmaps typically exhibit more background pixels than 2D heatmaps, they are especially challenging to use as training data. Furthermore, 3D heatmaps incur a higher computational cost to represent each landmark heatmap independently. To address these challenges, we represented the 3D heatmap within a singular 3D space~\cite{zhang2019adversarial} and trained the network using the Adaptive Wing (AWing) loss~\cite{wang2019adaptive}, which improves foreground pixel training. Thus, the pre-stage loss can be obtained as follows:

\begin{equation}
    \mathcal{L}_{pre} = {\rm AWing}(H, \hat{H}) + \frac{1}{3} ||p - \hat{p}||_{2}^{2},
\label{eq:loss_prestage}
\end{equation}

\noindent where AWing  denotes the 3D heatmap regression loss function; $\hat{H}$ and $H$ represent 3D heatmaps predicted from the 3D heatmap regression network and ground truth, respectively; and $\hat{p}$ and $p$ are the Euler angles of the head pose predicted from the pre-pose regression network and ground truth, respectively. The hyperparameters of the AWing loss function are the same as those defined in Wang et al.~\cite{wang2019adaptive}. The ground truth Euler angles for the pre-pose were calculated as the angular differences between the mean and sample landmarks. To quantify the similarity between the two landmark sets, we employed the least-squares approach, with the scale factor $s$ calculated as follows:

\begin{equation}
    s = \sqrt{\frac{\sum_{i=0}^{N-1}(\mathrm{x}_{k, i} - \bar{\mathrm{x}}_{k})^2}{\sum_{i=0}^{N-1}(\mathrm{x}_{m, i} - \bar{\mathrm{x}}_{m})^2}},
\label{eq:scale}
\end{equation}

\noindent where $\mathrm{x}_{k}$ and $\mathrm{x}_{m}$ are the sample and mean keypoint landmarks, respectively, and $\bar{\rm{x}}_{k}$ and $\bar{\rm{x}}_{m}$ are the centroids corresponding to said landmarks.

\begin{equation}
    M = (\mathrm{x}_{k}-\bar{\mathrm{x}}_{k})(\mathrm{x}_{m}-\bar{\mathrm{x}}_{m})^T,
\label{eq:cov}
\end{equation}

\noindent The covariance matrix $M$ is calculated using Equation~(\ref{eq:cov}). The rotation matrix is calculated by $[U,\Sigma,V] = \mathrm{SVD}(M)$ and Equation~(\ref{eq:rotmat})

\begin{equation}
    R = UV^T,
\label{eq:rotmat}
\end{equation}

\noindent The rotation matrix $R$ is transformed into Euler angles $p$ for pre-pose training. The translation can be calculated using the scale factor and rotation matrix:

\begin{equation}
    T = \mathrm{x}_{k} - sR * \mathrm{x}_{m},
\label{eq:trans}
\end{equation}

We adopted the feature extraction and 3D heatmap regression network structures proposed by Wang et al.~\cite{wang2019adaptive}, with the residual block in the latter replaced by the PFRB described in Section~\ref{sec:pfrb}. For downsampling, we employed average pooling throughout all networks except the first convolutional layer, where we used a stride of 2. For the feature extraction and pre-pose regression networks, we applied SE~\cite{hu2018squeeze} without pre-pose. 

\subsection{Dual-dimensional network}
\label{sec:ddn}

\begin{figure}[!htb]
	\centering
		\includegraphics[width=135mm]{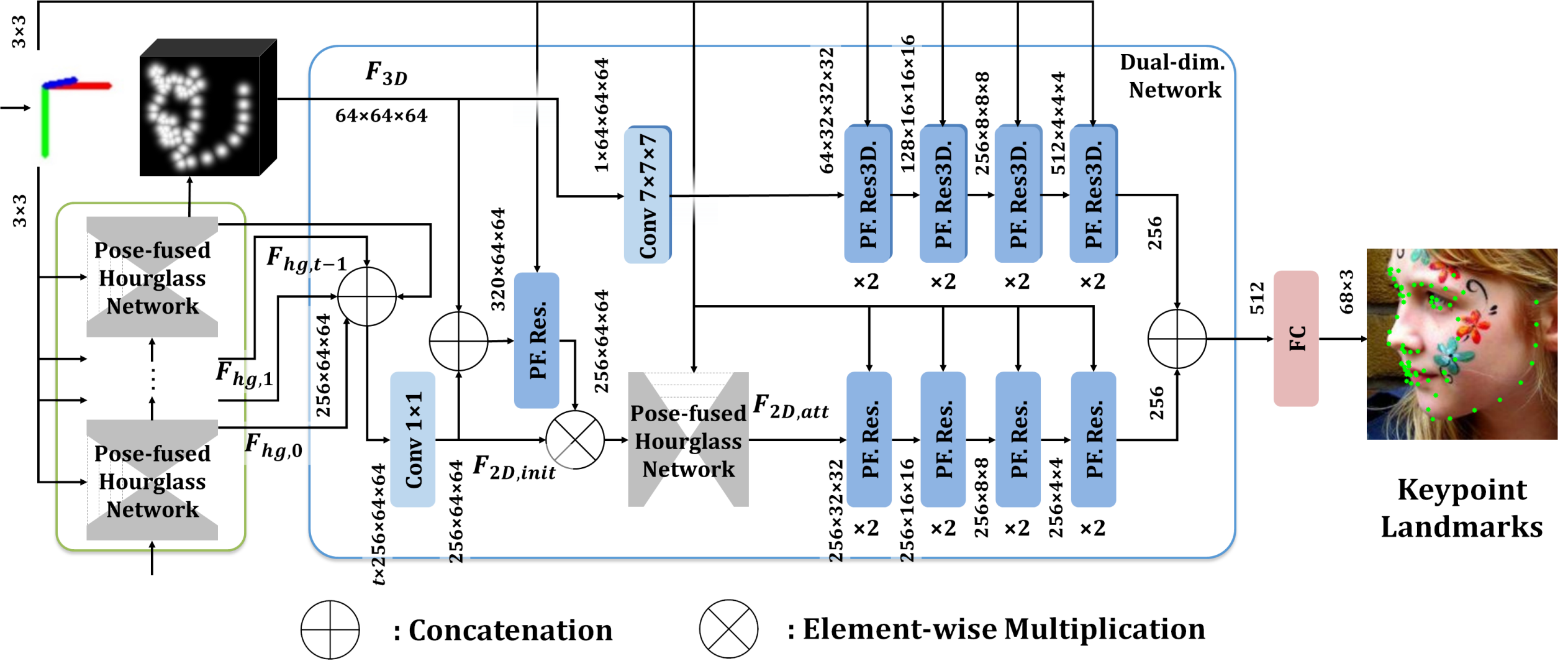}
	  \caption{Structure of dual-dimensional network.}
	  \label{fig:ddn}
\end{figure}

It is difficult to directly transform the heatmap represented in a single 3D space into the coordinates of facial landmarks, as the 3D space includes the heatmap of each landmark. Zhang et al.~\cite{zhang2019adversarial} transformed 3D coordinates through a subnetwork using a single 3D heatmap as a feature map, extending upon previous studies~\cite{pavlakos2017coarse, jackson2017large} using volumetric data. Although these methods are effective for predicting 3D coordinates using a single 3D heatmap, their results are highly dependent upon heatmap quality. Accordingly, we designed a dual-dimensional network that uses 2D feature maps extracted from the 3D heatmap regression network along with the 3D heatmap, thereby utilizing diverse feature maps. Figure~\ref{fig:ddn} illustrates the structure of the proposed dual-dimensional network. The layers of each dimension are designed with a structure that iteratively performs two residual block operations and downsampling, and the feature vectors of each dimension are concatenated before propagating to the fully-connected layer. The 2D feature maps are the last feature maps of each hourglass network in the 3D heatmap regression network. The early low-level feature maps exhibit noise and unrefined rich information, whereas the latter high-level feature maps exhibit less noise but also significant semantic information for the 3D heatmap regression task. To optimize feature diversity and thereby improve performance, these features can be combined into multilevel feature maps. Accordingly, we concatenated multilevel 2D feature maps with each level feature map in Figure~\ref{fig:ddn}. These 2D feature maps are reduced via a $1\times1$ convolutional layer, as follows:

\begin{equation}
    F_{2D} = \mathcal{F}_{1\times1}(F_{hg,0} \oplus F_{hg,1} \oplus ... \oplus F_{hg, t-1}),
\label{eq:multilvfeat2d}
\end{equation}

\noindent where $\mathcal{F}_{1\times1}$ is a $1\times1$ convolution, $F_{hg,t}$ represent single-level feature maps from the hourglass network, and $t$ is an index representing the hourglass network level. The 2D feature maps are concatenated with the predicted 3D heatmap using the depth dimensions as channels, with weights calculated by the PFRB for spatial attention. The final 2D feature maps are obtained by a single hourglass network following multiplication by the weights:

\begin{equation}
    F_{2D,att} = \mathcal{F}_{hgatt}(F_{2D} \otimes \mathcal{F}_{pfres}(F_{2D} \oplus F_{3D})),
\label{eq:feat2datt}
\end{equation}

\noindent Here, $\mathcal{F}_{hgatt}$ is the hourglass network, $\mathcal{F}_{pfres}$ is the PFRB, $\otimes$ denotes element-wise multiplication, and $F_{3D}$ is the 3D heatmap. Because $F_{3D}$ spans a single 3D space, it has one channel before being passed to the dual-dimensional network. This channel is increased by the first $7\times7\times7$ convolutional layer of the dual-dimensional network. The PFRB for the 3D convolution displaces a $3\times3\times3$ kernel. We trained the dual-dimensional network using the mean-squared loss for sparse face alignment:

\begin{equation}
    \mathcal{L}_{k} = \frac{1}{n_{k}}||\mathrm{x}_{k}-\mathrm{\hat{x}}_{k}||_{2}^{2},
\label{eq:loss_poststage}
\end{equation}

\noindent where $n_{k}$ is the number of keypoint landmarks and $\mathrm{\hat{x}}_{k}$ is the predicted keypoint landmarks. 

\subsection{Dense face alignment training based on knowledge distillation}

\begin{figure}[!htb]
	\centering
		\includegraphics[width=135mm]{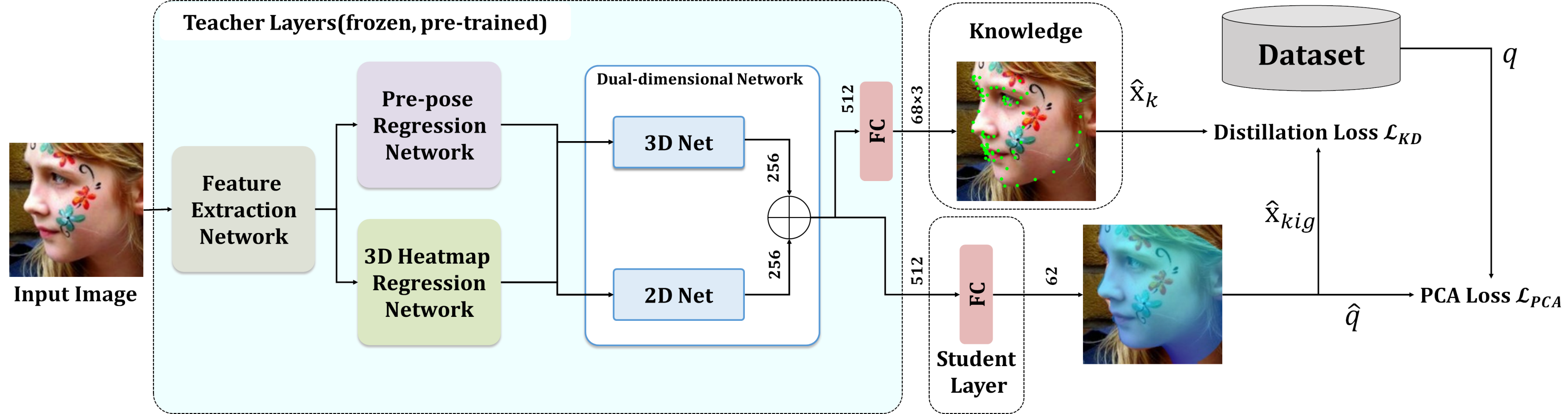}
	  \caption{Flow of dense face alignment training based on knowledge distillation. The fully-connected layer added to the end of the dual-dimensional network is trained alone. The loss function used for dense face alignment training uses the predicted keypoint landmarks.}
	  \label{fig:densefa}
\end{figure}

Prior 3D face alignment studies~\cite{zhu2016face, guo2020towards, wu2021synergy, li2023dsfnet, feng2018joint, ruan2021sadrnet} largely employed dense face alignment to predict facial geometry. 3DMM-based dense face alignment is the task of predicting geometric landmarks, including keypoint landmarks. Although geometric landmarks offer the advantage of being able to represent detailed facial information, they are more difficult to train than keypoint landmarks owing to the relatively large number of prediction targets. Previous studies~\cite{zhu2016face, guo2020towards, wu2021synergy} used the PCA parameters of the 3DMM to improve performance by defining loss functions that consider errors between the keypoint landmarks and predicted geometric landmarks. To increase the effectiveness of keypoint landmarks during training, we used a network trained for sparse face alignment. Figure~\ref{fig:densefa} illustrates the flow of dense face alignment training, which occurs in the fully-connected layer appended to the trained dual-dimensional network. The layer parameters of the trained dual-dimensional network are frozen to maintain the semantic information and performance of sparse face alignment. The output of the fully-connected layer is the PCA parameter defined as 

\begin{equation}
    q = [R, T,\alpha_{id}, \alpha_{exp}],
\label{eq:pcaparam}
\end{equation}

\noindent where $R$ is a scaled vectorized rotation matrix, $T$ is a 3D translation vector, $\alpha_{id}$ is a 3D shape parameter, and $\alpha_{exp}$ is the expression parameter of the 3DMM. As in Guo et al.~\cite{guo2020towards}, the PCA parameter $q$ has 62 dimensions, with nine dimensions of $R$, three dimensions of $T$, 40 dimensions of $\alpha_{id}$, and 10 dimensions of $\alpha_{exp}$. Furthermore, $q$ is calculated by z-score normalization $q = (q - \mu_{q} )⁄\sigma_{q}$  to minimize variability, where $\mu_{q}$ and $\sigma_{q}$ are the mean and standard deviation of the sample parameters. 

\begin{equation}
    \mathrm{x}_{g} = R * (\mathrm{\bar{x}}_{g} + \mathrm{A}_{id} \mathrm{\alpha}_{id} + \mathrm{A}_{exp} \mathrm{\alpha}_{exp}) + T,
\label{eq:pcabackproj}
\end{equation}

\noindent The geometric landmark $\mathrm{x}_{g}$ in Equation~(\ref{eq:pcabackproj}) is calculated using $q$ and the other parameters, where $\mathrm{\bar{x}}_{g}$ is the mean geometric landmark, and $\mathrm{A}_{id}$ and $\mathrm{A}_{exp}$ are the shape and expression bases of the 3DMM, respectively. The predicted keypoint landmarks are defined as knowledge and distilled into the fully-connected layer for training.

Knowledge distillation has been primarily used to improve the performance of small networks using knowledge from large networks~\cite{gou2021knowledge}. In the present study, we propose K2G which is a method using knowledge from sparse face alignment to train for dense face alignment. The loss function used for dense face alignment training is defined as follows:

\begin{equation}
    \mathcal{L}_{d} = \mathcal{L}_{PCA} + \mathcal{L}_{KD} = \frac{1}{n_{q}} ||q - \hat{q}||_{2}^{2} + \frac{1}{n_{k}} || \mathrm{\hat{x}}_{k} - \mathrm{\hat{x}}_{kig}||_{2}^{2},
\label{eq:denseloss}
\end{equation}

\noindent where $n_{q}$ is the number of PCA parameters, $\mathcal{L}_{PCA}$  is the loss function of the PCA parameters, $\mathcal{L}_{KD}$  is the loss function of knowledge distillation, $q$ is the ground truth of the PCA parameters, $\hat{q}$ is the predicted PCA parameters, and $\mathrm{\hat{x}}_{kig}$ is the keypoint landmark within the predicted geometric landmarks. The predicted keypoint landmarks can be used for geometric landmark training, as they are a subset of the set of geometric landmarks. Although this training approach yields an improvement in performance, the accuracy of predicted geometric landmarks is insufficient compared to that of predicted keypoint landmarks. To alleviate this problem, we propose G2K which is a method using the predicted geometry landmarks are aligned to the predicted keypoint landmarks. This alignment method transforms geometric landmarks using the similarity of both keypoint landmark sets, as shown in Equations~(\ref{eq:scale})-(\ref{eq:trans}).

Additionally, small PFA networks can be trained using knowledge distillation. Using the network defined in Sections~\ref{sec:prenet} and~\ref{sec:ddn} (denoted as PFA-L) as a basis, we designed smaller networks denoted as PFA-M and PFA-S, and trained them using keypoint landmarks predicted using PFA-L. Table~\ref{tbl:pfanet} lists the characteristics of the PFA structures, showing differences in terms of input resolution, channels, and residual blocks.

\begin{table}[!htb]
\caption{PFA network structures classified by layer type and number of output channels. R2D is a residual block~\cite{bulat2017binarized} with SE~\cite{hu2018squeeze}, FC is a fully-connected layer, PHG is a pose-fused hourglass network, Att. is a spatial attention, D is depth, and P2D and P3D are the pose-fused residual blocks of the 2D and 3D kernels, respectively.}\label{tbl:pfanet}

\scalebox{0.63}{
\renewcommand{\arraystretch}{1.2} 
\centering
\begin{tabular}{cccccccc}
\toprule
 \multicolumn{2}{c}{Model} &\multicolumn{2}{c}{PFA-L} & \multicolumn{2}{c}{PFA-M} & \multicolumn{2}{c}{PFA-S}\\
\midrule
 \multicolumn{2}{c}{Input Size} & \multicolumn{2}{c}{$3\times256\times256$} & \multicolumn{2}{c}{$3\times64\times64$} & \multicolumn{2}{c}{$3\times64\times64$}\\
\midrule
 \multirow{10}{*}{\parbox{1cm}{\centering Pre-\\stage}} & \multirow{3}{*}{\parbox{2cm}{\centering Feature\\Extraction}} & \multicolumn{2}{c}{Conv $7\times7$, 64} & \multicolumn{2}{c}{Conv $3\times3$, 64} & \multicolumn{2}{c}{Conv $3\times3$, 64}\\
 & & \multicolumn{2}{c}{R2D, 128} & \multicolumn{2}{c}{R2D, 128} & \multicolumn{2}{c}{R2D, 64}\\
 & & \multicolumn{2}{c}{R2D, 128} & \multicolumn{2}{c}{R2D, 128} & \multicolumn{2}{c}{R2D, 64}\\
\cline{2-8}
 & \multirow{5}{*}{\parbox{2cm}{\centering Pre-pose\\Regression}} & \multicolumn{2}{c}{R2D, 256} & \multicolumn{2}{c}{R2D, 256} & \multicolumn{2}{c}{R2D, 128}\\
 & & \multicolumn{2}{c}{R2D, 512} & \multicolumn{2}{c}{R2D, 512} & \multicolumn{2}{c}{R2D, 256}\\
 & & \multicolumn{2}{c}{R2D, 512} & \multicolumn{2}{c}{R2D, 512} & \multicolumn{2}{c}{R2D, 256}\\
 & & \multicolumn{2}{c}{R2D, 512} & \multicolumn{2}{c}{FC, 3} & \multicolumn{2}{c}{FC, 3}\\
 & & \multicolumn{2}{c}{FC, 3} & \multicolumn{2}{c}{-} & \multicolumn{2}{c}{-}\\
\cline{2-8}
 &\parbox{2cm} {\vspace{2mm} \begin{spacing}{1.1} \centering 3D Heatmap\\ Regression \end{spacing}\vspace{2mm}} & \multicolumn{2}{c}{\parbox{2.5cm}{\begin{spacing}{1.1} \centering 4-stacked\\PHGs(D=4),\\$64\times64\times64$ \end{spacing}}} & \multicolumn{2}{c}{\parbox{2.5cm}{ \begin{spacing}{1.1} \centering 2-stacked\\PHGs(D=2),\\$16\times16\times16$ \end{spacing}}} &  \multicolumn{2}{c}{\parbox{2.5cm}{\begin{spacing}{1.1} \centering 2-stacked\\PHGs(D=2),\\$16\times16\times16$\end{spacing}}}\\
\midrule
 \multirow{7}{*}{\parbox{1cm}{\centering Post-\\stage}} & \multirow{5}{*}{\parbox{2cm}{\centering Dual-\\dimensional}} & \parbox{2.5cm}{\begin{spacing}{1} \centering PHG Att.\\(D=4), 256 \end{spacing}\vspace{2mm}} & \parbox{2.5cm}{\begin{spacing}{1} \centering Conv\\$7\times7\times7$, 64\end{spacing}\vspace{2mm}} & \parbox{2.5cm}{\begin{spacing}{1} \centering PHG Att.\\(D=2), 256\end{spacing}\vspace{2mm}} & \parbox{2.5cm}{\begin{spacing}{1} \centering Conv\\$3\times3\times3$, 64\end{spacing}\vspace{2mm}} & \parbox{2.5cm}{\begin{spacing}{1} \centering PHG Att.\\(D=2), 128\end{spacing}\vspace{2mm}} & \parbox{2.5cm}{\begin{spacing}{1} \centering Conv\\ $3\times3\times3$, 64\end{spacing}\vspace{2mm}}\\
 & & P2D$\times2$, 256 & P3D$\times2$, 128 & P2D, 256 & P3D, 128 & P2D, 128 & P3D, 128\\
 & & P2D$\times2$, 256 & P3D$\times2$, 256 & P2D, 256 & P3D, 256 & P2D, 128 & P3D, 128\\
 & & P2D$\times2$, 256 & P3D$\times2$, 512 & P2D, 256 & P3D, 256 & - & -\\
 & & P2D$\times2$, 256 & P3D$\times2$, 256 & - & - & - & -\\
\cline{2-8}
 & Sparse FA & \multicolumn{2}{c}{FC, $68\times3$} & \multicolumn{2}{c}{FC, $68\times3$} & \multicolumn{2}{c}{FC, $68\times3$}\\
 & Dense FA & \multicolumn{2}{c}{FC, 62} & \multicolumn{2}{c}{FC, 62} & \multicolumn{2}{c}{FC, 62}\\
\bottomrule
\end{tabular}
}
\end{table}

\section{Experiments}

We trained our PFAs on the 300W-LP dataset~\cite{zhu2016face}, a synthetic version of the 300W dataset~\cite{sagonas2013300} comprising samples of various head poses, and subsequently evaluated them on the AFLW2000-3D~\cite{zhu2016face} and AFLW~\cite{koestinger2011annotated} datasets for face alignment. In addition, we evaluated the PFA-L-H network designed for head pose estimation on the AFLW2000-3D and BIWI datasets~\cite{fanelli2013random}. All PFAs were trained from scratch.

\subsection{Implementation details}
We cropped facial images from the 300W-LP, AFLW2000-3D, and AFLW datasets using the bounding boxes provided by the datasets. Because the BIWI dataset does not provide a bounding box, we used the MTCNN~\cite{zhang2016joint} face detector to crop the images according to the evaluation protocol defined by Yang et al.~\cite{yang2019fsa}. However, the bounding box generated by MTCNN exhibited a domain gap with the ground truth bounding box specified for the 300W-LP dataset. Therefore, we initially used MTCNN for cropping, and then predicted keypoint landmarks using PFA-L. Subsequently, we cropped the facial images again using bounding boxes around the predicted keypoint landmarks.

The pre-pose regression network predicts the pitch, yaw, and roll angles in radians, whereas the 3D heatmap regression network predicts the 3D heatmap represented in a single space. Unlike the network structure proposed by Zhang et al.~\cite{zhang2019adversarial}, only the last hourglass network in the 3D heatmap regression network predicts the 3D heatmap. The residual block comprising the hourglass network, which is based on a hierarchical, parallel \text{\&} MS block~\cite{bulat2017binarized}.

We augmented the data by applying sequential transformations including ±15\text{\%} random scaling, rotation, translation of ±25 pixels, 50\text{\%} flipping, and 0–50\text{\%} occlusion~\cite{park2021acn}. Rotation was randomly applied on only the roll axis, with a range of [-180°,180° ] in frontal images and [-90°,90° ] in profile images. We then randomly applied one of the following transformations: adding random Gaussian noise or JPEG compression artifacts; blending with grayscale; contrast, color, or lighting adjustment; power-law transformation; histogram equalization; or identity transformation.

We used the SGD optimizer with a batch size of 10, momentum of 0.9 and weight decay of $5\times10^{-4}$. The initial learning rate was set to $1\times10^{-2}$ in all training phases except that for dense face alignment, wherein it was set to $1\times10^{-6}$. The learning rate decayed by a factor of 0.1 every 15 epochs in the baseline training phases without knowledge. When training was performed based on knowledge distillation, the learning rate decayed by a factor of 0.9 whenever the average training loss value for the current epoch exceeded those for the previous 10 epochs. Baseline training was conducted for the pre-stage in all models as well as the sparse face alignment of PFA-L, whereas knowledge distillation was employed for all other training phases. Models were trained over 1,000 epochs for sparse face alignment and 500 epochs for dense face alignment. To evaluate head pose estimation performance, we trained PFA-L-H using the same environment as PFA-L with the omission of random rotations during augmentation.

\subsection{Face alignment}
We used the normalized mean error (NME) to evaluate face alignment performance. The NME, which represents the distance between predicted facial landmarks and ground truth, is expressed as follows:

\begin{equation}
    \mathrm{NME} = \frac{1}{N}\sum_{i=0}^{N-1}\frac{||\mathrm{\hat{x}}_{i} - \mathrm{x}_{i}||_2}{d},
\label{eq:nme}
\end{equation}

\noindent where $N$ is the number of landmarks, $\mathrm{\hat{x}}_{i}$ is the $i$-th predicted landmark, $\mathrm{x}_{i}$ represents the corresponding ground truth landmark, and $d$ is the normalization factor defined as the square root of the area in the bounding box. We also evaluated the dense face alignment performance of PFA-L in 3D face reconstruction with $d$ representing the interocular distance, i.e., the distance between the outer endpoints of the eyes. Test samples were divided among three balanced subsets corresponding to angular ranges of [0°, 30°), [30°, 60°), and [60°, 90°], as described by Zhu et al.~\cite{zhu2016face}. The NME of 3D face reconstruction was calculated after aligning the predicted geometric landmarks to the ground truth using the iterative closest point (ICP) algorithm, as described in previous studies~\cite{feng2018joint, ruan2021sadrnet}.

\subsubsection{Evaluation on AFLW2000-3D}
The AFLW2000-3D dataset, comprising the first 2,000 samples of the AFLW dataset, is the most widely-used dataset for 3D face alignment. The performance of sparse face alignment was evaluated using keypoint landmarks, whereas that of dense face alignment was evaluated using geometric landmarks with the neck and ears omitted.

\begin{table}[!htb]
\caption{Evaluation of face alignment on AFLW2000-3D dataset.}\label{tbl:aflw2000}
\scalebox{0.7}{
\centering
\begin{tabular}{ccccccccc}
\toprule
\multirow{3}{*}{Method} & \multicolumn{6}{c}{68 Keypoints} & \multicolumn{2}{c}{45K Geometry}\\
\cline{2-9}
 & \multicolumn{5}{c}{2D} & 3D & 2D & 3D\\
\cline{2-9}
 & [0°,30°) & [30°, 60°) & [60°, 90°] & Mean(A) & Mean(B) & Mean(A) & Mean(A) & Mean(A)\\
\midrule
3DDFA~\cite{zhu2016face} & 3.78 & 4.54 & 7.93 & 6.03 & 5.42 & 7.50 & 5.06 & 6.55\\
3D-FAN~\cite{bulat2017far} & 3.16 & 3.53 & 4.60 & - & 3.76 & - & - & -\\
PRN~\cite{feng2018joint} & 2.75 & 3.51 & 4.61 & 3.26 & 3.62 & 4.70 & 3.17 & 4.40\\
JVCR~\cite{zhang2019adversarial} & 2.69 & 3.08 & 4.15 & - & 3.31 & - & - & -\\
2DASL~\cite{tu20203d} & 2.75 & 3.44 & 4.41 & - & 3.53 & - & - & -\\
3DDFAv2~\cite{guo2020towards} & 2.63 & 3.42 & 4.48 & - & 3.51 & - & - & 4.18\\
SADRNet~\cite{ruan2021sadrnet} & 2.66 & 3.30 & 4.42 & 3.05 & 3.46 & 4.33 & 2.93 & 4.02\\
SynergyNet~\cite{wu2021synergy} & 2.65 & 3.30 & 4.27 & 3.03 & 3.41 & - & - & -\\
DSFNet-f~\cite{li2023dsfnet} & 2.46 & 3.20 & 4.16 & - & 3.27 & - & - & 3.80\\
DSLPT~\cite{xia2023robust} & 2.51 & 3.40 & 4.32 & - & 3.41 & - & - & -\\
\midrule
PFA-L(Ours) & 2.47 & 3.13 & 4.03 & 2.83 & 3.21 & 4.13 & 2.74 & 3.84\\
PFA-M(Ours) & 2.54 & 3.15 & 4.03 & 2.88 & 3.24 & 4.27 & 2.77 & 3.89\\
PFA-S(Ours) & 2.64 & 3.23 & 4.10 & 2.98 & 3.33 & 4.30 & 2.84 & 3.98\\
\bottomrule
\end{tabular}
}
\end{table}

\begin{figure}[!htb]
	\centering
		\includegraphics[width=135mm]{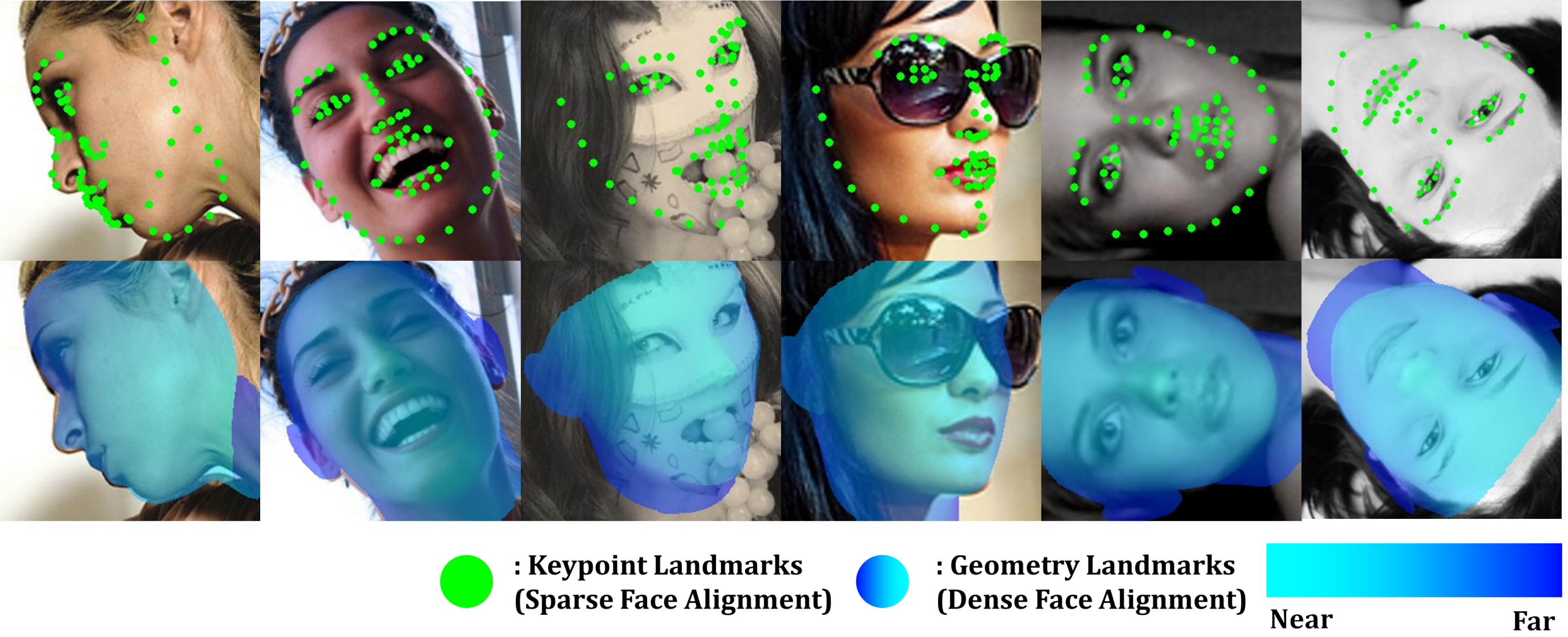}
	  \caption{Visualized examples of PFA-L output on AFLW2000-3D dataset. The green points in the first row denote the results of sparse face alignment, whereas the blue areas in the second row denote the results of dense face alignment. Darker shades of blue indicate a further distance.}
	  \label{fig:aflw2000}
\end{figure}

Table~\ref{tbl:aflw2000} and Figure~\ref{fig:aflw2000} present the PFA results for the AFLW2000-3D dataset. In Table~\ref{tbl:aflw2000}, Mean(A) is the mean value of all samples and Mean(B) is the mean value of the balanced subsets. The proposed PFAs exhibited competitive performance. Specifically, PFA-L obtained improvements of -0.41\%, 2.19\%, 3.13\%, 1.83\%, and -1.05\% for the respective categories of [0°, 30°), [30°, 60°), [60°, 90°], Mean(B) of the 2D keypoint landmarks, and Mean(A) of the 3D geometric landmarks, compared to DSFNet-f~\cite{li2023dsfnet}. Furthermore, PFA-L exhibited improvements of 4.62\% and 4.48\% in the Mean(A) values of the 3D keypoint landmarks and the 2D geometric landmarks compared to SADRNet~\cite{ruan2021sadrnet}.

\begin{table}[!htb]
\caption{Evaluation on re-annotated version of AFLW2000-3D dataset for face alignment.}\label{tbl:aflw2000re}
\scalebox{1.0}{
\centering
\begin{tabular}{cccccc}
\toprule
\multirow{2}{*}{Method} & \multicolumn{5}{c}{68 2D Keytpoints, Re-annotated}\\
\cline{2-6}
 & [0°,30°) & [30°, 60°) & [60°, 90°] & Mean(A) & Mean(B)\\
\midrule
3DDFA~\cite{zhu2016face} & 2.84 & 3.52 & 5.15 & -- & 3.83\\
PRN~\cite{feng2018joint} & 2.35 & 2.78 & 4.22 & - & 3.11\\
MGCNet~\cite{shang2020self} & 2.72 & 3.12 & 3.76 & - & 3.20\\
3DDFAv2~\cite{guo2020towards} & 2.84 & 3.03 & 4.13 & - & 3.33\\
SynergyNet~\cite{wu2021synergy} & 2.05 & 2.49 & 3.52 & - & 2.65\\
\midrule
PFA-L(Ours) & 1.90 & 2.09 & 3.11 & 2.12 & 2.37\\
PFA-M(Ours) & 2.06 & 2.03 & 3.25 & 2.29 & 2.54\\
PFA-S(Ours) & 2.17 & 2.46 & 3.45 & 2.42 & 2.70\\
\bottomrule
\end{tabular}
}
\end{table}

Bluat et al.~\cite{bulat2017far} determined an issue in the quality of ground truth landmarks in the AFLW2000-3D dataset, and provided re-annotated 2D keypoint landmarks for evaluation. Table~\ref{tbl:aflw2000re} presents our evaluation results using the re-annotated landmarks, with PFA-L demonstrating improvements of 7.32\%, 16.06\%, 11.65\%, and 10.57\% in the [0°, 30°), [30°, 60°), [60°, 90°], and Mean(B) categories compared to SynergyNet~\cite{wu2021synergy}. 

\subsection{Evaluation on AFLW}

\begin{figure}[!htb]
	\centering
		\includegraphics[width=135mm]{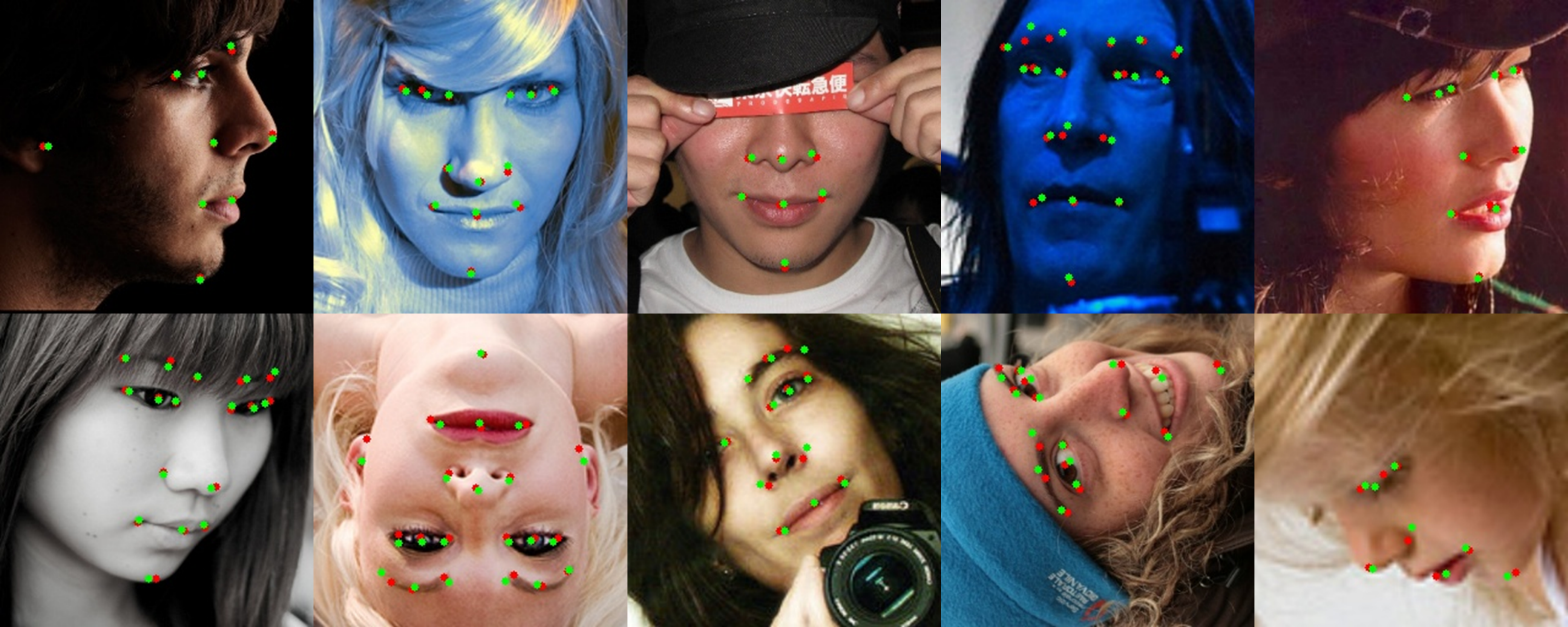}
	  \caption{Visualized output of PFA-L on AFLW dataset. Red points denote the ground truth, whereas green dots denote predicted landmarks. Occluded landmarks are not shown.}
	  \label{fig:aflw}
\end{figure}

The AFLW dataset~\cite{koestinger2011annotated} largely comprises posed faces. Guo et al.~\cite{guo2020towards} evaluated their method using 21 visible 2D keypoint landmarks from the AFLW dataset. Using the same evaluation protocol, we compared the performance of the proposed method with that of a state-of-the-art method. The results in Table~\ref{tbl:aflw} demonstrate that PFA-L obtained improvements of 9.31\%, 8.42\%, 10.27\%, and 9.36\% in the [0°, 30°), [30°, 60°), [60°, 90°], and Mean(B) categories compared to SynergyNet~\cite{wu2021synergy}. The results of this experiment are visualized in Figure~\ref{fig:aflw}, demonstrating the robustness of PFA to large poses and occlusions.

\begin{table}[!htb]
\caption{Evaluation on face alignment on AFLW dataset.}\label{tbl:aflw}
\scalebox{1.0}{
\centering
\begin{tabular}{cccccc}
\toprule
\multirow{2}{*}{Method} & \multicolumn{5}{c}{21 visible 2D Keypoints}\\
\cline{2-6}
 & [0°,30°) & [30°, 60°) & [60°, 90°] & Mean(A) & Mean(B)\\
\midrule
3DDFA~\cite{zhu2016face} & 4.75 & 4.83 & 6.39 & - & 5.32\\
3D-FAN~\cite{bulat2017far} & 4.40 & 4.52 & 5.17 & - & 4.69\\
PRN~\cite{feng2018joint} & 4.19 & 4.69 & 5.45 & - & 4.77\\
3DDFAv2~\cite{guo2020towards} & 3.98 & 4.31 & 4.99 & - & 4.43\\
SynergyNet~\cite{wu2021synergy} & 3.76 & 3.92 & 4.48 & - & 4.06\\
\midrule
PFA-L(Ours) & 3.41 & 3.59 & 4.02 & 3.58 & 3.68\\
PFA-M(Ours) & 3.48 & 3.65 & 4.15 & 3.65 & 3.76\\
PFA-S(Ours) & 3.57 & 3.72 & 4.25 & 3.74 & 3.85\\
\bottomrule
\end{tabular}
}
\end{table}

\subsubsection{3D face reconstruction}
In the task of 3D face reconstruction, the accuracy of 3D geometry must be prioritized over landmark locations. We assessed the accuracy of geometric shapes predicted by PFA-L in the dense face alignment task by evaluating 3D face reconstruction. The results listed in Table~\ref{tbl:aflw2000reconst} demonstrate improvements of 0.95\%, 1.46\%, 1.79\%, and 1.23\% in the [0°, 30°), [30°, 60°), [60°, 90°], and Mean(A) categories between PFA-L and SADRNet~\cite{ruan2021sadrnet}.

\begin{table}[!htb]
\caption{Evaluation of 3D face reconstruction on AFLW2000-3D dataset.}\label{tbl:aflw2000reconst}
\scalebox{1.0}{
\centering
\begin{tabular}{cccccc}
\toprule
Method & [0°,30°) & [30°, 60°) & [60°, 90°] & Mean(A) & Mean(B)\\
\midrule
3DDFA~\cite{zhu2016face} & - & - & - & 5.36 & -\\
DeFA~\cite{liu2017dense} & - & - & - & 5.64 & -\\
PRN~\cite{feng2018joint} & 3.72 & 4.04 & 4.45 & 3.96 & -\\
SPDT~\cite{piao2019semi} & - & - & - & 3.70 & -\\
SADRNet~\cite{ruan2021sadrnet} & 3.17 & 3.42 & 3.36 & 3.25 & -\\
DSFNet-is~\cite{li2023dsfnet} & - & - & - & 3.19 & -\\
\midrule
PFA-L(Ours) & 3.14 & 3.37 & 3.30 & 3.21 & 3.27\\
PFA-M(Ours) & 3.20 & 3.45 & 3.50 & 3.29 & 3.38\\
PFA-S(Ours) & 3.20 & 3.48 & 3.62 & 3.32 & 3.44\\
\bottomrule
\end{tabular}
}
\end{table}

Although PFA exhibited superior overall performance in the face alignment task compared to existing methods, the degree of improvement was insignificant for 3D face reconstruction. The scale, rotation, and translation errors were reduced because the predicted geometric landmarks were aligned with the ground truth landmarks using the ICP algorithm prior to evaluation. Consequently, head pose information has a more significant impact on the accuracy of face alignment than that of 3D face reconstruction.

\subsection{Head pose estimation}
\begin{table}[!htb]
\caption{Evaluation of head pose estimation on AFLW2000-3D dataset.}\label{tbl:pose}
\scalebox{1.0}{
\centering
\begin{tabular}{ccccccccc}
\toprule
\multirow{2}{*}{Method} & \multicolumn{4}{c}{AFLW2000-3D} & \multicolumn{4}{c}{BIWI}\\
\cline{2-9}
 & Yaw & Pitch & Roll & Mean & Yaw & Pitch & Roll & Mean\\
\midrule
HopeNet~\cite{ruiz2018fine} & 6.47 & 6.56 & 5.44 & 6.16 & 4.81 & 6.61 & 3.27 & 4.90\\
FSA-Net~\cite{yang2019fsa} & 6.50 & 6.08 & 4.64 & 5.07 & 4.27 & 4.96 & 2.76 & 4.00\\
QuatNet~\cite{hsu2018quatnet} & 3.91 & 5.62 & 3.92 & 4.50 & 2.94 & 5.49 & 4.01 & 4.15\\
HPE~\cite{huang2020improving} & 4.80 & 6.18 & 4.87 & 5.28 & 3.12 & 5.18 & 4.57 & 4.29\\
WHENet-V~\cite{zhou2020whenet} & 4.44 & 5.75 & 4.31 & 4.83 & 3.60 & 4.10 & 2.73 & 3.48\\
TriNet~\cite{cao2021vector} & 4.04 & 5.77 & 4.20 & 4.67 & 4.11 & 4.76 & 3.05 & 3.97\\
MNN~\cite{valle2020multi} & 3.34 & 4.69 & 3.48 & 3.83 & 3.98 & 4.61 & 2.39 & 3.66\\
FDN~\cite{zhang2020fdn} & 3.78 & 5.61 & 3.88 & 4.42 & 4.52 & 4.70 & 2.56 & 3.93\\
SADRNet~\cite{ruan2021sadrnet} & 2.93 & 4.43 & 2.95 & 3.44 & - & - & - & -\\
SynergyNet~\cite{wu2021synergy} & 3.42 & 4.09 & 2.55 & 3.35 & - & - & - & -\\
6DRepNet~\cite{hempel20226d} & 3.63 & 4.91 & 3.37 & 3.97 & 3.24 & 4.48 & 2.68 & 3.47\\
DAD-3DNet~\cite{martyniuk2022dad} & 3.08 & 4.76 & 3.15 & 3.66 & - & - & - & -\\
TokenHPE~\cite{zhang2023tokenhpe} & 4.36 & 5.54 & 4.08 & 4.66 & 3.95 & 4.51 & 2.71 & 3.72\\
DSFNet-f~\cite{li2023dsfnet} & 2.65 & 4.28 & 2.82 & 3.25 & - & - & - & -\\
\midrule
PFA-L-H(Ours) & 2.33 & 4.07 & 2.85 & 3.08 & 3.15 & 4.43 & 2.56 & 3.38\\
\bottomrule
\end{tabular}
}
\end{table}

\begin{figure}[!htb]
	\centering
		\includegraphics[width=135mm]{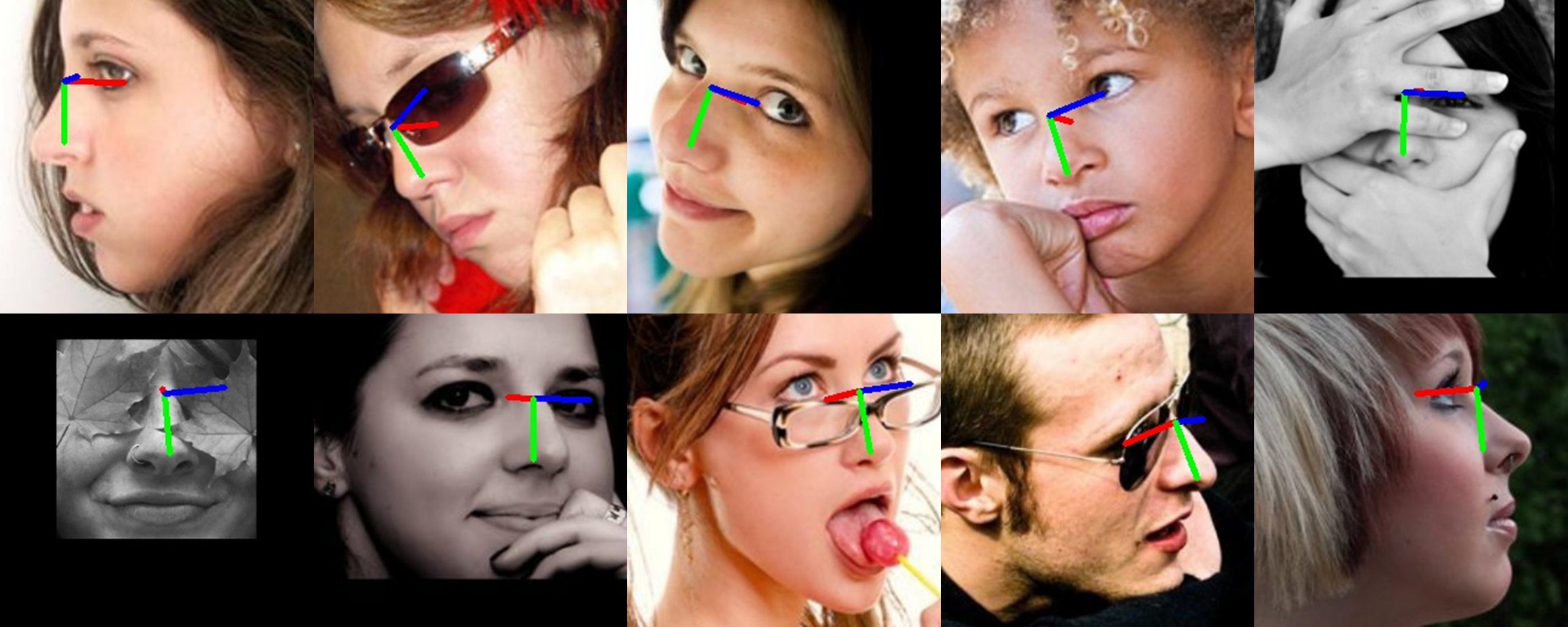}
	  \caption{Visualized output of PFA-L-H on AFLW2000-3D dataset. Green, blue and, red lines denote the yaw, pitch, and roll axes, respectively.}
	  \label{fig:aflw2000pose}
\end{figure}

Although PFA was designed for face alignment, head pose estimation can also be performed by setting the dual-dimensional network to output Euler angles in place of keypoint landmarks, with the modified network denoted as PFA-L-H. Accordingly, the pre-pose is refined into the face orientation provided by the dataset. We conducted a performance comparison between PFA and existing state-of-the-art methods on the AFLW2000-3D and BIWI datasets. Following previous studies~\cite{yang2019fsa, cao2021vector, ruiz2018fine}, we only considered samples with rotation angles in the range of [-99°, 99°], omitting the 31 samples in AFLW2000-3D falling outside this range. Because the BIWI dataset does not provide facial landmarks, we did not evaluate using another protocol which is used the BIWI dataset for training.

Figure~\ref{fig:aflw2000pose} and Table~\ref{tbl:pose} present the results of this evaluation. We observe that PFA-L-H obtained improvements of 31.87\%, 0.49\%, -11.76\%, and 8.06\% in yaw, pitch, roll and mean compared to SynergyNet [8] on the AFLW2000-3D dataset, and 2.78\%, 1.12\%, 4.48\%, and 2.59\% compared to 6DRepNet~\cite{hempel20226d} on the BIWI dataset. Thus, PFA-L-H exhibited the best overall performance.

\subsection{Model complexity}
\begin{figure}[!htb]
\subfloat[AFLW2000-3D]{{\includegraphics[width=0.5\textwidth ]{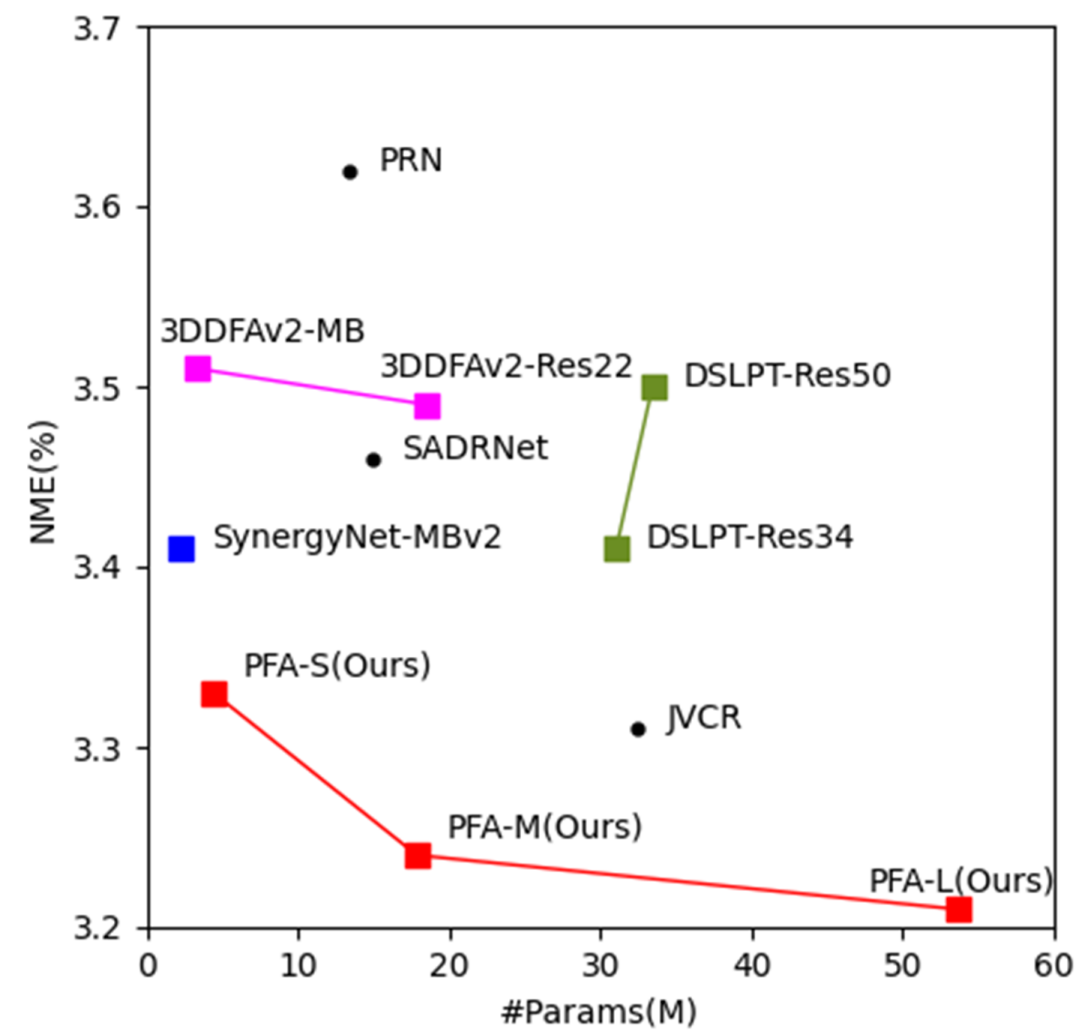} }}
\subfloat[AFLW]{{\includegraphics[width=0.5\textwidth ]{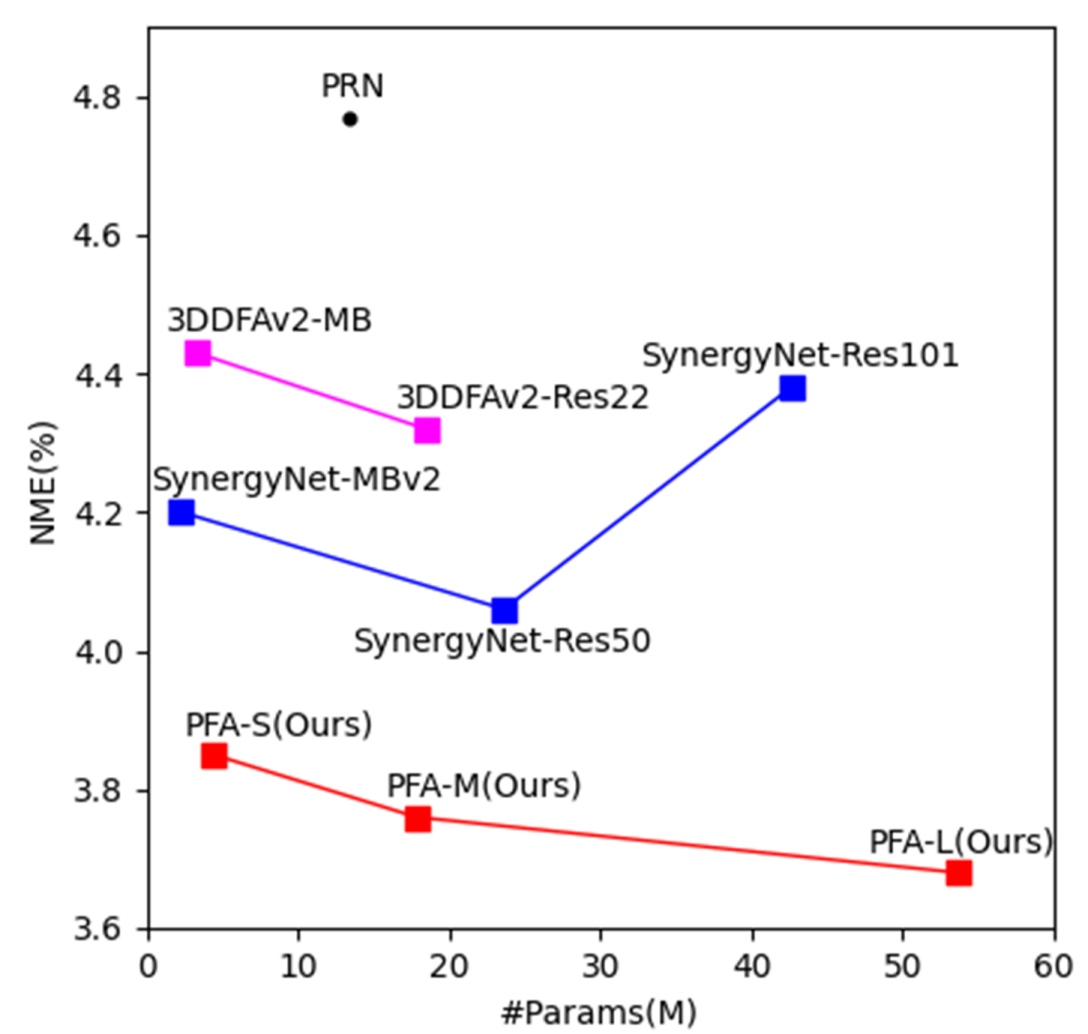} }}
\caption{Number of parameters versus NME. (a) and (b) depict evaluations on the AFLW2000-3D and AFLW datasets, respectively. MB denotes a MobileNet~\cite{howard2017mobilenets} backbone network, MBv2 represents a MobileNetv2~\cite{sandler2018mobilenetv2} backbone network, and Res refers to a ResNet~\cite{he2016deep} backbone network.}
\label{fig:complexity}
\end{figure}

\begin{table}[!htb]
\caption{Comparison of model complexity.}\label{tbl:complexity}
\scalebox{0.90}{
\centering
\begin{tabular}{cccccc}
\toprule
\multicolumn{2}{c}{\multirow{2}{*}{Method}} & \multicolumn{2}{c}{68 , 2D NME Mean(B)} & \multirow{2}{*}{Params(M)} & \multirow{2}{*}{FLOPs(G)}\\
\cline{3-4}
 & & AFLW2000-3D & AFLW & &\\
\midrule
\multicolumn{2}{c}{PRN~\cite{feng2018joint}} & 3.62 & 4.77 & 13.40 & 6.19\\
\multicolumn{2}{c}{JVCR~\cite{zhang2019adversarial}} & 3.31 & - & 32.47 & 19.42\\
\multirow{2}{*}{3DDFAv2~\cite{guo2020towards}} & MobileNet & 3.51 & 4.43 & 3.27 & 0.18\\
 & ResNet22 & 3.49 & 4.32 & 18.45 & 2.66\\
\multicolumn{2}{c}{SADRNet~\cite{ruan2021sadrnet}} & 3.46 & - & 14.88 & 10.37\\
\multirow{3}{*}{SynergyNet~\cite{wu2021synergy}} & MobileNetv2 & 3.41 & 4.20 & 2.30 & 0.10\\
 & ResNet50 & - & 4.06 & 23.64 & 1.27\\
 & ResNet101 & - & 4.38 & 42.63 & 2.49\\
\multirow{2}{*}{DSLPT~\cite{xia2023robust}} & ResNet34 & 3.41 & - & 31.05 & 7.02\\
 & ResNet50 & 3.50 & - & 33.46 & 7.69\\
\midrule
\multicolumn{2}{c}{PFA-L(Ours)} & 3.21 & 3.68 & 53.64 & 73.06\\
\multicolumn{2}{c}{PFA-M(Ours)} & 3.24 & 3.76 & 17.88 & 1.80\\
\multicolumn{2}{c}{PFA-S(Ours)} & 3.33 & 3.85 & 4.44 & 0.64\\
\bottomrule
\end{tabular}
}
\end{table}

We compared the proposed and state-of-the-art methods in terms of complexity using the NME of the 2D keypoint landmarks, number of network parameters, and FLOPs. The results of this comparison are presented in Figure~\ref{fig:complexity} and Table~\ref{tbl:complexity}. In the figure, the larger models – namely 3DDFAv2~\cite{guo2020towards}, SynergyNet ~\cite{wu2021synergy}, and DSLPT~\cite{xia2023robust} – exhibit only marginal differences in performance compared to PFA. Furthermore, PFA-L exhibits efficient performance despite its large network size. The smaller networks trained through knowledge distillation – namely PFA-M and PFA-S – exhibit minimal performance degradation with significantly less parameters. Specifically, they exhibited performance decreases of 0.93\% and 3.74\%, respectively, on the ALFW2000-3D dataset compared to PFA-L, despite the 0.33- and 0.08-fold decreases in the number of parameters and 0.02- and 0.01-fold decreases in FLOPs, respectively.

\subsection{Ablation Study}
The proposed PFA performs face alignment by fusing predicted head pose information into feature maps. Furthermore, we implemented dual-dimensional regression to use 2D and 3D feature maps simultaneously, and formulated a training method based on knowledge distillation for dense face alignment. To verify the individual effectiveness of these contributions, we conducted an ablation study on the AFLW2000-3D dataset. First, we compared performance according to the quality of the pre-pose input to PFA-L. Subsequently, we evaluated the performance of PFA-L when the dual-dimensional network was replaced with a single-dimension network using only single-dimension kernels. Finally, we evaluated the dense face alignment method based on knowledge distillation.

\subsubsection{Effectiveness of pre-pose}
To improve performance, the predicted pre-pose is passed as input to the PFRB. We evaluated the performance of PFA while varying the pre-pose quality to validate the effectiveness of the pre-pose. Residual blocks within the network without a prepose were applied from the fusion of the pre-pose to the general channel attention~\cite{hu2018squeeze}.

\begin{table}[!htb]
\caption{Performance evaluation with variable pre-pose quality.}\label{tbl:posequality}
\centering
\begin{tabular}{ccccc}
\toprule
\multicolumn{2}{c}{Component} & \multicolumn{3}{c}{Choice}\\
\midrule
\multirow{2}{*}{Pre-pose} & Ground truth & - & - &\checkmark\\
 & Predicted & - & \checkmark & -\\
\midrule
\multirow{2}{*}{\parbox{4cm}{\centering NME\\(68 keypoints)}} & 2D Mean(B) & 3.33 & 3.21 & 2.28\\
 & 3D Mean(A) & 4.22 & 4.13 & 2.74\\
\bottomrule
\end{tabular}
\end{table}

Table~\ref{tbl:posequality} presents the results of this evaluation. The network trained without the pre-pose exhibited the worst performance, whereas that trained with the ground truth pre-pose exhibited the best performance. The predicted and ground truth pre-poses improved performance by 3.60\% and 31.53\% in terms of 2D NME Mean(B), respectively, compared to the case without the pre-pose. This improvement indicates that head pose information significantly contributes to face alignment performance.

\subsubsection{Effectiveness of dual-dimensional network in coordinate regression}
PFA performs face alignment through a dual-dimensional network using 2D and 3D feature maps in the post-stage. The 2D multilevel feature maps extracted by the four-stacked hourglass networks in the pre-stage are 2D feature maps, while the predicted 3D heatmap encompasses the 3D feature maps. Spatial attention was applied to the 2D feature maps using the 3D heatmap.

\begin{table}[!htb]
\caption{Evaluation of feature map types in dual-dimensional network.}\label{tbl:dimtest}
\scalebox{0.88}{
\centering
\begin{tabular}{cccccc}
\toprule
Feature Map Type & 2D & 3D & 2D(3DAtt.) & 2D $\oplus$ 3D & 2D(3DAtt.) $\oplus$ 3D\\
\midrule
NME(68, 2D Mean(B)) & 3.28 & 3.32 & 3.27 & 3.24 & 3.21\\
\bottomrule
\end{tabular}
}
\end{table}

Table~\ref{tbl:dimtest} presents the evaluation results obtained using identical pre-stage networks while training the post-stage networks according to feature map type. The dual-dimensional network exhibited an improvement of 2.41\% compared to the network using the 3D heatmap. Furthermore, the application of spatial attention to the 2D feature maps improved performance by 0.93\%. Overall, performance improved by 3.31\%.

\subsubsection{Effectiveness of training based on knowledge distillation}
The dense face alignment training process employs knowledge distillation using keypoint landmarks predicted by the dual-dimensional network. We compared a baseline training phase that uses ground truth keypoint landmarks to the proposed knowledge-distillation-based process, K2G. In addition, we evaluated G2K which is the alignment method of predicted geometric landmarks to more accurately predicted keypoint landmarks. According to the results shown in Table~\ref{tbl:kddense}, K2G exhibited a slight improvement; however, the combination of G2K and K2G improved performance by 1.29\% compared to the baseline.

\begin{table}[!htb]
\caption{Effectiveness of training based on knowledge distillation and alignment using predicted keypoint landmarks.}\label{tbl:kddense}
\scalebox{0.88}{
\centering
\begin{tabular}{cccc}
\toprule
Method & Baseline & K2G(Training) & K2G(Training)+G2K(Aligning)\\
\midrule
NME(45K, 3D Mean(A)) & 3.89 & 3.88 & 3.84\\
\bottomrule
\end{tabular}
}
\end{table}

Table~\ref{tbl:kddense} compares results between the baseline using the loss function (refer to Equation~(\ref{eq:loss_poststage})) and knowledge distillation using predicted keypoint landmarks. The latter method exhibited improvements of 5.26\% and 4.58\% on the AFLW2000-3D dataset and 3.59\% and 4.70\% on the AFLW dataset, in PFA-M and PFA-S, respectively.

\begin{table}[!htb]
\caption{Effectiveness of training based on knowledge distillation and alignment using predicted keypoint landmarks.}\label{tbl:kddense}
\centering
\begin{tabular}{ccccc}
\toprule
\multirow{2}{*}{Model} & \multicolumn{2}{c}{AFLW2000-3D} & \multicolumn{2}{c}{AFLW}\\
\cline{2-5}
 & Baseline & KD & Baseline & KD\\
\midrule
PFA-L & 3.21 & - & 3.68 & -\\
PFA-M & 3.42 & 3.24 & 3.90 & 3.76\\
PFA-S & 3.49 & 3.33 & 4.04 & 3.85\\
\bottomrule
\end{tabular}
\end{table}

\section{Conclusion}
In this paper, we propose the pose-fused face alignment (PFA) model, wherein head pose information is fused into feature maps within a residual block to enhance the accuracy of face alignment. We designed the PFA process with two stages: the pre-stage, wherein features are extracted, and the post-stage, where said features are used to perform face alignment. In the pre-stage, head pose information is extracted along with 2D feature maps and a 3D heatmap. These features are then used in the post-stage to perform face alignment through a dual-dimensional network. Our experimental results reveal that head pose information is closely linked to the accuracy of face alignment as a significant contributing factor. Improvements in performance were also obtained by simultaneously using the 2D feature maps and 3D heatmap in the post-stage, as well as performing training based on knowledge distillation. However, the proposed method did not significantly improve 3D face reconstruction compared to state-of-the-art methods. Accordingly, we plan to improve the accuracy of 3D face reconstruction by considering facial expression information, which is an important facial property, are alongside head pose information.


\end{document}